\documentclass[lettersize,journal]{IEEEtran}
\usepackage{amsmath,amsfonts}
\usepackage{algorithmic}
\usepackage{algorithm}
\usepackage{array}
\usepackage[caption=false,font=normalsize,labelfont=sf,textfont=sf]{subfig}
\usepackage{textcomp}
\usepackage{stfloats}
\usepackage{url}
\usepackage{verbatim}
\usepackage{graphicx}
\usepackage{cite}

\usepackage{multirow}
\usepackage{color}
\usepackage{booktabs}

\begin{document}

\title{Uncertainty-Driven Multi-Scale Feature Fusion \\ Network 
      for Real-time Image Deraining}

\author{Ming Tong, Xuefeng Yan, Yongzhen Wang
\thanks{M. Tong and Y. Wang are with the School of Computer Science and Technology, Nanjing University of Aeronautics and Astronautics, Nanjing 210016, China (e-mail:tongming@nuaa.edu.cn, wangyz@nuaa.edu.cn).}
\thanks{X. Yan is with the School of Computer Science and Technology, Nanjing University of Aeronautics and Astronautics, Nanjing 210016, China, also with the Collaborative Innovation Center of Novel Software Technology and Industrialization, Nanjing 210093, China (e-mail: yxf@nuaa.edu.cn).}
}

\markboth{Journal of \LaTeX\ Class Files,~Vol.~14, No.~8, August~2021}%
{Shell \MakeLowercase{\textit{et al.}}: A Sample Article Using IEEEtran.cls for IEEE Journals}


\maketitle

\begin{abstract}
Visual-based measurement systems are frequently affected by rainy weather due to the degradation caused by rain streaks in captured images, and existing imaging devices struggle to address this issue in real-time. While most efforts leverage deep networks for image deraining and have made progress, their large parameter sizes hinder deployment on resource-constrained devices. Additionally, these data-driven models often produce deterministic results, without considering their inherent epistemic uncertainty, which can lead to undesired reconstruction errors. Well-calibrated uncertainty can help alleviate prediction errors and assist measurement devices in mitigating risks and improving usability. Therefore, we propose an Uncertainty-Driven Multi-Scale Feature Fusion Network (UMFFNet) that learns the probability mapping distribution between paired images to estimate uncertainty. Specifically, we introduce an uncertainty feature fusion block (UFFB) that utilizes uncertainty information to dynamically enhance acquired features and focus on blurry regions obscured by rain streaks, reducing prediction errors. In addition, to further boost the performance of UMFFNet, we fused feature information from multiple scales to guide the network for efficient collaborative rain removal. Extensive experiments demonstrate that UMFFNet achieves significant performance improvements with few parameters, surpassing state-of-the-art image deraining methods.
\end{abstract}



\section{Introduction}
\IEEEPARstart{O}{utdoor} measurement systems often necessitate non-corrupted images to ensure optimal performance in various applications, such as autonomous driving\cite{cai2021yolov4}, surveillance\cite{cao2020haze,wang2020real}, and remote sensing\cite{lu2021attention}. However, images captured during rainy conditions are always compromised, leading to information loss due to the obscuration of rain streaks. So, finding an effective and reliable way to eliminate rain streaks from images and restore the background is essential for vision-based measurement applications.

Previous image deraining methods can be categorized into two main types: model-based and learning-based methods.  Early model-based methods primarily rely on statistical analysis, such as sparse coding\cite{kang2011automatic} and Gaussian mixture models\cite{li2016rain}, to analyze rain-free backgrounds and rain streaks. But these methods only capture shallow information,  whereas the background content displays diversity, which makes it challenging to find appropriate priors. As a result, this challenge can lead to inaccurate deraining outcomes.

\begin{figure}[!t]
\centering
\includegraphics[width=3.5in]{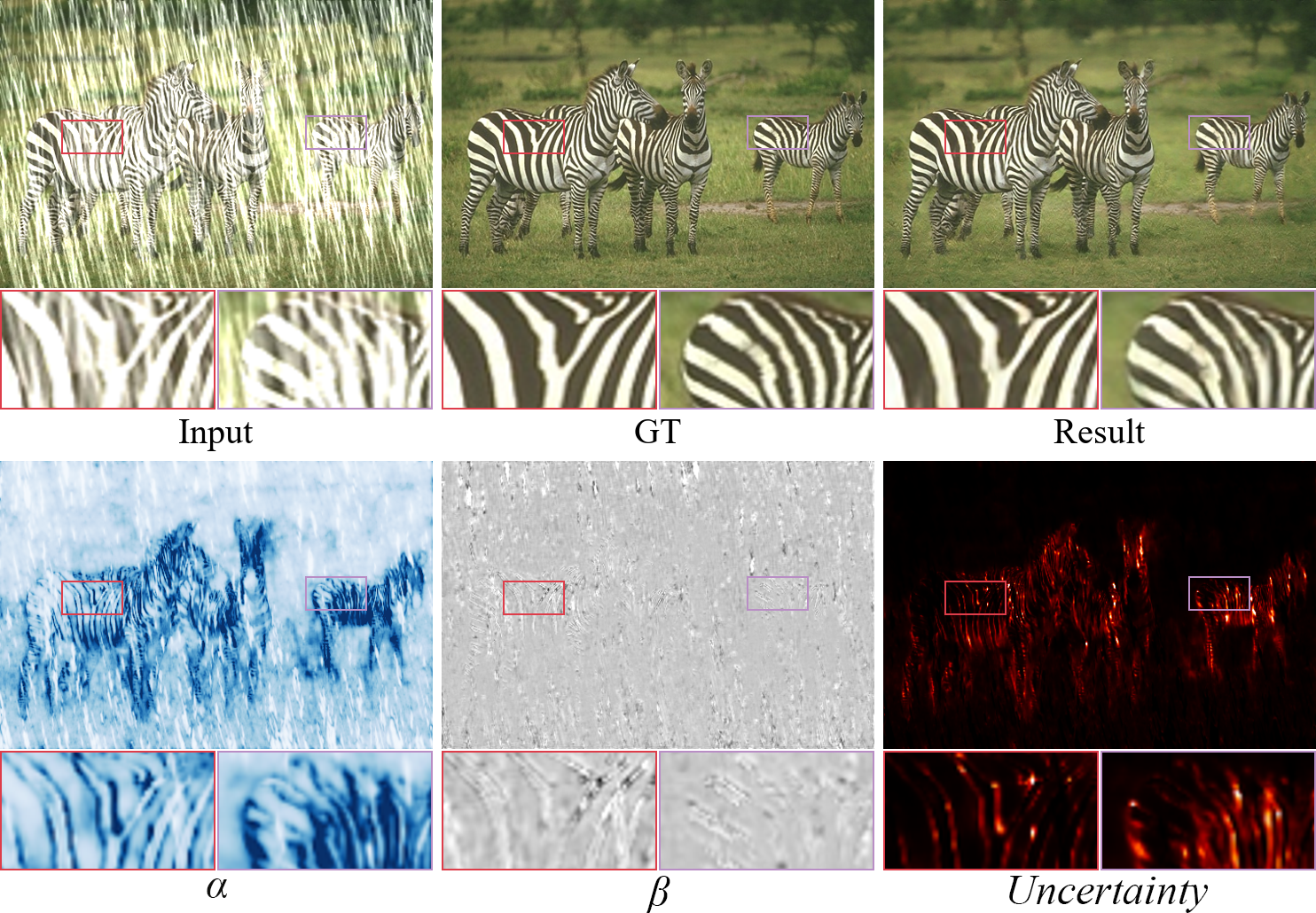}
\caption{In the second row, the scale ($\alpha$) and shape ($\beta$) represent spatial parameters generated by the model, which are used for uncertainty estimation. It can be observed that a strong correlation exists between the $Uncertainty$ and the image reconstruction error. As the reconstruction error increases, the uncertainty also increases. In other words, when the model fails to accurately recover the background texture and edge details of the image, it leads to higher uncertainty. Therefore, the uncertainty information can guide the model in enhancing the pixel representation of regions with significant reconstruction errors.
}
\label{fig_1}
\end{figure}

Recently, learning-based methods such as Convolutional Neural Networks (CNNs) and Transformers have gained significant popularity in image deraining tasks due to their superior ability to represent rain streak shapes and background object textures/edges. These deep learning methods capture the complex mappings from rain images to clean images by extracting hierarchical features. Additionally, rain-related assumptions/constraints, such as background features\cite{fan2018residual}, binary rain mask\cite{yang2017deep}, and rain density\cite{li2017single}, are incorporated into the networks to enhance their understanding of rain characteristics, facilitating the effective removal of rain layers and restoration of rain-free images. In addition to rain-related constraints, deraining networks incorporate elaborately designed complex blocks to enhance their feature learning capabilities. Zhang et al.\cite{zhang2018density} introduced a Multi-stream dense network that integrates dense blocks and CNNs to jointly estimate rain density and remove rain. Yang et al.\cite{yang2019scale} proposed a dilated residual dense network that combines dense blocks and residual networks, delivering exceptional performance in heavy rain conditions. Ren et al.\cite{ren2019progressive} presented a progressive recurrent network for effective deraining, using recursive aggregation of residual blocks and LSTM connections across different recursions. Recent several Transformer-based architectures, including Uformer\cite{wang2022uformer}, Restormer\cite{zamir2022restormer}, and SwinIR\cite{liang2021swinir}, have showcased outstanding performance in computer vision tasks. However, these self-attention-based networks may impose a substantial computational burden, potentially leading to excessive computation and resource consumption during training. Essentially, the performance improvements of these methods mainly arise from their intricate model designs. Nevertheless, as the depth and width of the networks increase, the number of model parameters and system complexity also continue to rise, with only modest improvements in performance.

Moreover, despite the significant advancements made by these methods, the majority of them only focus on learning the deterministic one-to-one mapping between rainy and non-rainy images, neglecting the consideration of uncertainty in the results. The quality of results obtained from tests and measurements is crucial in practical vision-based measurement applications such as autonomous driving and medical imaging, and estimating uncertainty provides an effective means to assess outcome quality\cite{al2021machine,da2012measurement,ferrero2006fully}. Well-calibrated uncertainty enables experts or devices to intervene in highly uncertain predictions, thereby enhancing decision-making reliability\cite{upadhyay2022bayescap,schwarting2018planning,van2021artificial}. Therefore, researchers are encouraged to address the uncertainty in deep learning in order to enhance the reliability of measurement systems based on deep learning\cite{al2021machine}. In depth estimation, Lucas et al.\cite{teixeira2020aerial} utilized uncertainty estimation to eliminate erroneous depth values and achieve high-precision predictions. In medical image denoising tasks, \cite{laves2020uncertainty} applies Bayesian methods with Monte Carlo dropout to provide reliable uncertainty estimates for solving the issues of hallucinations and artifacts. \cite{yasarla2019uncertainty} and \cite{chen2021joint} have mentioned that incorporating uncertainty estimation alongside pixel regression can offer more informed decisions and potentially enhance prediction accuracy. In visual tasks, typically, larger uncertainty values often correspond to greater reconstruction errors. However, how to utilize uncertainty estimation to bolster the reliability of deraining models is still under investigation.

Based on the above issues, we propose the following question: Can a CNN architecture be designed to yield considerable deraining results while minimizing computational costs and enhancing confidence level? 

The design of the network architecture and basic blocks plays a vital role in influencing the network's learning capacity and, consequently, its deraining performance. Therefore, we reviewed the key architecture design of image derain networks. Firstly, we observed that the multi-scale architecture is widely used and performs well in deraining tasks, such as the multi-scale progressive fusion network\cite{jiang2020multi}, PyramidDerain\cite{fu2019lightweight}, and uncertainty guided multi-scale attention network\cite{yasarla2019uncertainty}. This architecture facilitates the learning of distinct scale features of rain streaks, enabling the model to effectively remove multi-scale rain streaks. Building upon these state-of-the-art approaches, we employed a multi-scale UNet architecture that leverages simple connections to fuse feature information from various scales, which preserves contextual image details. Secondly, for the design of basic blocks, we opted for the less computationally demanding ResNet block\cite{he2016deep} instead of the more resource-intensive Transformer block\cite{liu2021swin}, thus mitigating computational costs. Furthermore, we integrated a simplified channel attention mechanism\cite{chen2022simple} into the basic block to augment the model's feature expression capabilities. 

Meanwhile, the inherent black-box nature and epistemic uncertainty of CNNs lead to deviations between the deraining results and the ground truth. The reconstructed results conform to the expectations of the deraining model based on previously learned information, but the training datasets typically do not encompass all possible scenarios. As a result, the reconstruction may contain false artifacts or dark noise. The noise embedded in the image not only affects the visual quality but also hinders clear differentiation of valid information, thereby impacting subsequent computer vision-based systems. Considering the high correlation between reconstruction errors and uncertainty, we introduce uncertainty estimation to improve the reliability and performance of the model. The network models the pixel residuals between the output and the ground truth using a generalized Gaussian distribution to provide uncertainty estimation. We then utilized pixel-level uncertainty maps to focus the network on regions with significant reconstruction errors (typically areas covered by rain streaks) in the rainy images, improving the representation of pixels in the uncertain regions and obtaining more convincing deraining results.


The main contributions of this paper are as follows:

\begin{itemize}
    \item Our proposed deraining network estimates uncertainty by using a generalized Gaussian distribution.  Based on the predicted uncertainty information, an uncertainty feature fusion block(UFFB) is designed to dynamically enhance local feature learning, which facilitates the model to generate high-quality, confident deraining images.
	\item We employ a multi-scale architecture to fully mine the similarities of multi-scale rain streaks. Additionally, we design three basic blocks, Residual Attention Block, Supervised Feature Fusion Block, and Multi-stage Fusion Block, to effectively extract and integrate contextual feature information from multiple scales for improving image deraining.
    \item Three variants of the proposed UMFFNet were trained using different depths. The experiments show that the tiny model UMFFNet-T achieves a PSNR index of 38.86 on the rain100L dataset with only 1.52M parameters. Furthermore, the deraining model with well-calibrated uncertainty instills sufficient confidence for future reliable applications of deep learning-based measurement instruments.
\end{itemize}

\section{Related Work}
\subsection{Single image deraining}
Image deraining aims to remove rain streaks and restore the original background scene from the degraded image. Early image deraining methods developed various priors to represent the properties of the rain and background layers. Lin et al. \cite{kang2011automatic} used dictionary learning and sparse coding to decompose the extracted high-frequency components of rainy images and then utilized it to eliminate sparse light rain streaks. Li et al. \cite{li2016rain} employed a Gaussian mixture model to separate the rain and background layers, and then removed small-scale rain streaks using total variation. With the rise of deep learning, learning-based methods, such as CNNs, recurrent neural networks, generative adversarial networks (GAN), and Transformer, have gained popularity and shown remarkable performance in deraining tasks. However, these networks often lack the ability to model uncertainty due to their deterministic one-to-one mapping approach. To enhance derained image prediction, we incorporate uncertainty estimation within the deraining network, enabling improved reconstruction of uncertain regions.

\subsection{Multi-Scale Approach}
Multi-scale methods\cite{jiang2020multi,fu2019lightweight,yasarla2019uncertainty,nah2017deep} leverage inputs and outputs at varying scales to enable networks to extract richer image information and effectively recover image details. PyramidDerain\cite{fu2019lightweight} combines Gaussian-Laplacian pyramid with deep neural networks, decomposing the deraining task into multiple sub-networks using the multi-scale technique for removing multi-scale rain streaks. Jiang et al.\cite{jiang2020multi} utilize a pyramid representation to jointly capture rain streaks of different scales, effectively extracting and integrating multi-scale information for progressive deraining. Apart from deraining, the multi-scale approach finds application in other image restoration tasks. Nah et al. \cite{nah2017deep} propose a multi-scale convolutional deblurring network that utilizes multi-scale blur kernel information to recover clear images from coarse to fine. In this paper, we combine the multi-scale approach with an encoder-decoder network that extracts features from images of different scales at each network layer and fuses them using simple connections to guide derain at different scales.

\subsection{Uncertainty Estimation}

Bayesian deep learning is a valuable approach for modeling uncertainty estimation. Uncertainty can be divided into two main categories: (1) epistemic uncertainty, arising from model parameters, and (2) aleatoric uncertainty, originating from inherent noise in measurement observations \cite{der2009aleatory}. In recent computer vision research, Bayesian modeling has been applied to capture uncertainty. For instance, in \cite{upadhyay2021uncertainty}, voxel-level uncertainty is quantified in predicted results to improve the output of undersampled MRI reconstruction task and modality propagation task, thereby offering a more reliable basis for clinical decision-making. Hong et al. \cite{hong2022uncertainty} utilize both aleatoric and epistemic uncertainty, predicted by the model, to adaptively enhance features learned through reinforcement learning and employ uncertainty-driven distillation loss to improve dehazing performance. In this paper, we adopt a heavy-tailed distribution to model the residual between derained and ground-truth images, enabling the estimation of uncertainty in predictions. We then leverage uncertainty representation to achieve more accurate deraining results.

\begin{figure*}[htb]
\begin{minipage}[b]{1.0\linewidth}
  \centering
  \centerline{\includegraphics[scale=0.9]{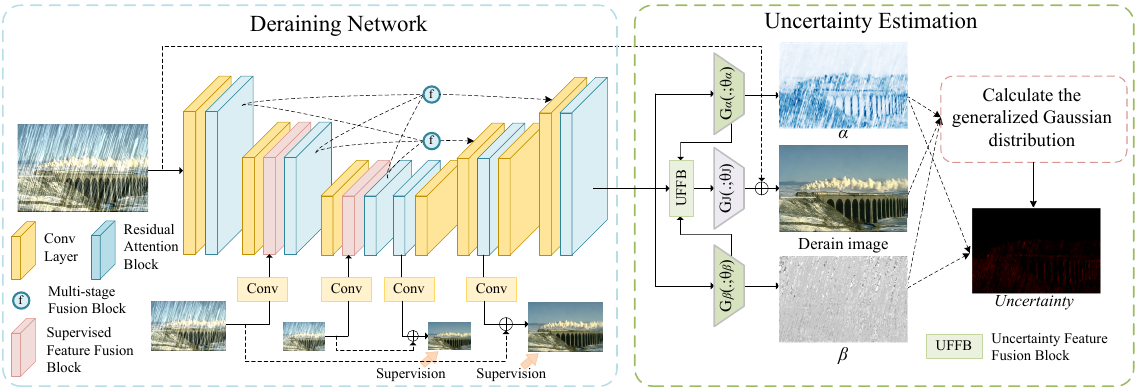}}
\end{minipage}
\caption{The overall architecture of Uncertainty-driven Multi-scale Feature Fusion Network (UMFFNet). UMFFNet comprises a deraining network and uncertainty estimation. The rainy images are downsampled to $1/2$ and $1/4$ scales, respectively, and combined with the original-scale image as inputs. The network generates deraining images at the corresponding scales. Uncertainty estimation is performed using the original-scale results. By leveraging the uncertainty maps, the network aims to minimize reconstruction errors and restore high-quality deraining images.}
\label{fig_2}
\end{figure*}
\section{Uncertainty-driven Multi-scale Feature Fusion Network}
Due to the latent similarity between images at different scales, we utilize a multi-scale structure to extract features across multiple scales and fuse the complementary feature information to acquire extensive contextual knowledge for joint deraining.    Additionally, considering that most existing deraining models do not fully mine the uncertain information and aiming to enhance the regions obscured by rain streaks, we model the pixel residuals using a generalized Gaussian distribution to estimate uncertainty.  Leveraging this uncertainty information, we guide the network optimization for deraining to achieve reliable and high-quality results. The overall architecture of UMFFNet is shown in Fig. \ref{fig_2}.

\subsection{Encoder-Decoder Deraining Network}
The network architecture of UMFFNet is mainly based on the simple U-Net\cite{ronneberger2015u}. The encoding and decoding parts comprise three different-scale encoder and decoder blocks. Given a rain image $I\in R^{H\times W\times 3}$, we downsample it to $1/2$ scale and $1/4$ scale to form the multi-scale inputs. The rain images at different scales are initially processed with $3\times3$ convolutions to extract shallow features, which are then fed into encoding blocks at various layers. Drawing inspiration from \cite{cho2021rethinking} and \cite{jiang2020multi}, we introduce a Supervised Feature Fusion Block between the two layers. This block fuses the scaled features from the previous layer and the shallow features of the downsampled rain images. These fused features are subsequently input into the encoding block to extract merged feature mappings from different pathways.  Similar to \cite{cho2021rethinking,mao2021deep}, we utilize a Multi-stage Fusion Block (MFB) to establish closer connections between contextual features across different scales. This block helps each layer in the decoder acquire multi-scale features and learn extensive contextual information, leading to improved deraining performance. Next, we will provide a detailed description of the specialized blocks used in the network, including RAB, SFFB, and MFB.extract the merged feature mappings from different pathways. Similar to \cite{cho2021rethinking,mao2021deep}, in order to establish closer connections between contextual features within the different scale, we employ a simple Multi-stage Fusion Block to connect information from different scales and pass it to the encoding block. Consequently, each sub-network in the decoder can obtain multi-scale features and learn extensive contextual information, thus enhancing the network's rain removal performance. Next, we will provide a detailed description of the special blocks used in the network, such as RAB, SFFB, and MFB.

\begin{figure*}[htb]
\begin{minipage}[b]{1.0\linewidth}
  \centering
  \centerline{\includegraphics[scale=1.1]{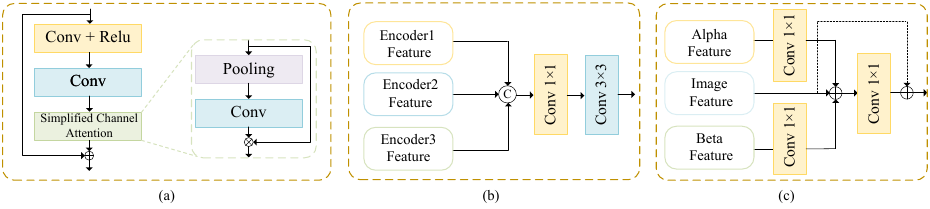}}
\end{minipage}
\caption{(a) Residual Attention Block. (b) Multi-stage Fusion Block. (c) Uncertainty Feature Fusion Block. }
\label{fig_3}
\end{figure*}

\textbf{Residual Attention Block(RAB)}. 
We employ Residual Attention Blocks as the building blocks of UMFFNet. RAB combines channel attention and residual block, as depicted in Figure \ref{fig_3} (a). The \cite{chen2022simple} finds that the network can be simplified by removing or replacing non-linear activation functions with multiplication. Based on this finding, we incorporate the simplified channel attention mechanism into the residual block, reducing the computational cost of network parameters while further enhancing the network's expressive capacity. Let us denote the input features as $x$. First, we apply a $3 \times 3$ convolution $Conv(.)$ and the ReLU function $\sigma (.)$ to process $x$, resulting in features $x_1$. Then, $x_1$ is fed into the attention block for global information aggregation and interaction. Finally, we achieve residual learning and compensation of distant information through skip connections \cite{deng2020detail}. The specific process is elucidated by the following equation:
\begin{equation}
\begin{aligned}
\label{eq1}
    x_1=conv(\sigma (conv(x)))
\end{aligned}
\end{equation}
\begin{equation}
\begin{aligned}
\label{eq2}
    out=x+x_1*conv(pool(x_1))
\end{aligned}
\end{equation}

\begin{figure}[htb]
\begin{minipage}[b]{1.0\linewidth}
  \centering
  \centerline{\includegraphics[width=7.5cm]{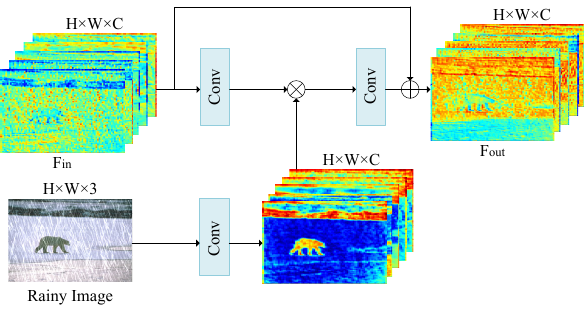}}
\end{minipage}
\caption{Supervised Feature Fusion Block.}
\label{fig_4}
\end{figure}

\textbf{Supervised Feature Fusion Block(SFFB)}. 
Recently, in the field of image restoration, the use of supervised attention modules between stages in multi-stage networks \cite{jiang2020multi} has led to significant performance improvements. In contrast to these multi-stage networks, we propose the Supervised Feature Fusion Block (SFFB) to incorporate valuable supervisory features from the original rainy images between each pair of encoders, helping the network supplement the lost original image information during the encoding process. Additionally, the network employs a gating mechanism to filter out low-information content features in the current encoding layer, allowing only effective features to pass through. The principle of SFFB is illustrated in Fig. \ref{fig_4}. The SFFB takes the input features $F_{in}\in R^{H\times W\times C} $ from the previous stage and the degraded rainy image $I\in R^{H\times W\times 3} $, first applies simple $1\times 1 $ convolutions and multiplication activation operations separately to obtain locally enhanced features $F_{in}^{'}$. Then, the enhanced features $F_{in}^{'}$ are added to the input features $F_{in}\in R^{H\times W\times C}$ after $1\times 1$ convolution to obtain the final enhanced feature representation $F_{out}$.

\textbf{Multi-stage Fusion Block(MFB)}. 
In our network, we introduce the MFB (see Fig. \ref{fig_3}(b)) between the encoder and the decoder. By concatenating the features from each encoder, we refine and aggregate them using $1\times1$ and $3\times3$ convolutions. The fused features are then fed to the decoder for further processing. The added MFB enhances the flow of feature information, allowing the network to effectively learn abundant contextual information and mitigate the impact of information loss.

\subsection{Uncertainty Estimation}
Let $D={(I_i,J_i)}^N_{i=1}$ denotes the training dataset from image domain $A$ and $B$ (i.e. $I_i \in A$, $J_i \in B$), where $(I_i,J_i)$ represents a pair of images, $I_i$ refers to the rainy input and $J_i$ refers to its Groud-truth. Motivated by previous works \cite{upadhyay2021uncertainty,upadhyay2021robustness}, we consider $\phi(.;\theta)$ parametrized by $\theta$ as a deep neural deraining network, which is designed to map images from the rainy set $A$ to the non-rainy set $B$. In this case, given an input $I_i$, the deraining model produces a pixel-wise prediction $\hat{J_i}=\phi(I_i:\theta)$ as the output. But the point output can not obtain the distribution of the prediction $P(J|I)$ and then the uncertainty in the output that is hard to capture. To leverage uncertainty to guide the network towards finding optimal $\theta^*$, the model needs to estimate the parameters to represent the distribution $P(J|I)$ and thus capture the uncertainty within the output, ultimately utilizing these parameters to maximize the likelihood function. Therefore, we treat the parameters $\theta$ of the deraining model $\phi(I;\theta): A\rightarrow B$ as a set of trainable parameters. That is, for a given rainy image input $I_i$, the model generates a set of parameter representations for the output:
\begin{equation}
\begin{aligned}
\label{eq3}
    \{\hat{J_i}, \rho_i, ..., \nu_i\}:=\phi(I_i:\theta)
\end{aligned}
\end{equation}
These parameters can be used to characterize the distribution, such that $P_{J|I}(J_i; {\hat{J_i}, \rho_i, ..., \nu_i})$. Then, the distribution $P(J|I)$ is selected to enable the use of closed-form solution $F$ dependent on the estimated parameters of the deraining model to predict uncertainty, while the likelihood function $L(\theta; D)$ is maximized to obtain the optimal parameter $ \theta^*$ of the network for achieving the best rain removal effect.
\begin{equation}
\begin{aligned}
\label{eq4}
 &   \theta^* :=   \mathop{argmax}\limits_{\theta}L(\theta;D) \\
 &   =\mathop{argmax}\limits_{\theta}\prod_{i=1}^{N}P_{J|I}(J_i;\{\hat{J_i}, \rho_i, ..., \nu_i\})
\end{aligned}
\end{equation}
\begin{equation}
\begin{aligned}
\label{eq5}
    Uncertainty(\hat{J}_i)=F( \rho_i, ..., \nu_i)
\end{aligned}
\end{equation}

In the case of distribution $P(J|I)$ in the above equation, recent studies have suggested that, in the presence of outliers and human factors, the residuals (the difference between predicted values and ground-truth values) of each pixel typically follow a heavy-tailed distribution \cite{upadhyay2021robustness,upadhyay2021uncertainty,upadhyay2022bayescap}. Therefore, to make our model robust to various outliers, we model the heteroscedastic generalized Gaussian distribution to simulate the heavy-tailed distribution of the output. When modeling the residuals using a heavy-tailed distribution, $\phi(I_i: \theta)$ is designed to predict the deraining image $\hat{J}_i$, scale $\alpha_i$ and shape $\beta_i$ as trainable parameters, i.e., $\{\hat{J}_i, \alpha_i, \beta_i\} := \phi(I_i: \theta)$. Consequently, the optimization problem described above can be formulated as follows:
\begin{equation}
\begin{aligned}
\label{eq6}
& \theta^* :=  \mathop{argmax}\limits_{\theta}L(\theta;D) \\
& =\mathop{argmax}\limits_{\theta}\prod_{i=1}^{N}\frac{\beta_i}{2\alpha_i\Gamma (\frac{1}{\beta_i})}{e^{-(|\hat{J}_i-J_i|/\alpha_i)}}^{\beta_i} \\
& = \mathop{argmin}\limits_{\theta}\sum_{i=1}^{N}(\frac{|\hat{J}_i-J_i|}{\alpha_i})^{\beta_i}-log\frac{\beta_i}{\alpha_i} + log\Gamma(\frac{1}{\beta_i})
\end{aligned}
\end{equation}
\begin{equation}
\begin{aligned}
\label{eq7}
 Uncertainty(\hat{J}_i)= \frac{\alpha_i^2\Gamma (\frac{3}{\beta_i})}{\Gamma(\frac{1}{\beta_i})}
\end{aligned}
\end{equation}
where $\Gamma(z)=\int_{0}^{\infty}x^{z-1}e^{-x}dx,\forall z>0$ express the Gamma function \cite{artin2015gamma}.
By leveraging these obtained parameters of different predictive distributions, we can calculate the uncertainty of the output. 

In the following sections, we will describe how to utilize the predicted uncertainty maps to guide the network to shift its attention towards uncertain pixels, aiming to enhance the deraining performance.

\textbf{Uncertainty Feature Fusion Block(UFFB)}. 
UFFB aims to improve derain performance by employing uncertain features. It can be directly observed that the estimated uncertainty values are correlated with the reconstruction error. Typically, the regions obscured by rain streaks exhibit higher levels of uncertainty. However, incorporating the predicted uncertain mappings, $\alpha_i$ and $\beta_i$, directly into the output rain-free image as conditional inputs may introduce undesirable interference due to the disparities between image features and uncertain mappings.

Therefore, we developed UFFB to assist the network in adaptively enhancing the learning of local features by leveraging uncertainty information, with the goal of accurately restoring regions that may have significant reconstruction errors.
The structure of UFFB is illustrated in Fig. \ref{fig_3}(c). The input of UFFB includes $\alpha_i$ and $\beta_i$ features from the uncertain estimation block, as well as image features from the UNet. The uncertain estimation block comprises three layers of $3\times3$ convolutions and ReLU activation functions, extracting features after the second convolutional and activation operations. These three types of features in UFFB are combined through summation, followed by refinement using a $1\times1$ convolution. The refined features are then added to the original image features to generate adaptively enhanced attention for further processing in the subsequent stage.

\subsection{Loss Function}
For model joint training, we utilized three distinct losses. We utilized the multi-scale content loss \cite{nah2017deep} $L_{con}$ as the primary loss function, and the multi-scale frequency reconstruction loss \cite{cho2021rethinking} $L_{fre}$ and uncertainty estimation loss $L_{ue}$ as auxiliary loss functions. Therefore, our overall loss function is expressed as follows:
\begin{equation}
\begin{aligned}
\label{eq8}
 L_{total}=L_{con}+ \lambda _1 L_{fre} +\lambda _2 L_{ue}
\end{aligned}
\end{equation}
where the weight parameters $\lambda _1$ and $\lambda _2$ are all set to 0.1.

The multi-scale approach aims to achieve derained output at each scale. Therefore, our network applies $L_1$ loss and frequency reconstruction loss to each layer. The frequency reconstruction loss is computed by applying a fast Fourier transform to the image signal and measuring the $L_1$ distance between the derained results and the ground truth image in the frequency domain:
\begin{equation}
\begin{aligned}
\label{eq9}
L_{con}=\sum_{k=1}^{K}||\hat{J}_k-J_k||_1
\end{aligned}
\end{equation}
where $\hat{J}_k$, $J_k$ represent the model result and ground-truth image at scale $k$, respectively.
\begin{equation}
\begin{aligned}
\label{eq10}
 L_{fre}=\sum_{k=1}^{K}||F_t(\hat{J}_k)-F_t(J_k)||_1
\end{aligned}
\end{equation}
where $F_t$ represents the fast Fourier transform.

As mentioned above, to estimate the uncertainty mapping, we minimize the negative log-likelihood for $\phi(.: \theta)$ to predict the parameters of the distribution. Equations (\ref{eq6}) and (\ref{eq7}) guide the network in estimating the distribution parameters and uncertainty. Hence, uncertainty estimation loss is actually the process of minimizing the negative log-likelihood for $\phi(.: \theta)$, which can be expressed as:
\begin{equation}
\begin{aligned}
\label{eq11}
L_{ue}=\sum_{i=1}^{N}(\frac{|\hat{J}_i-J_i|}{\alpha_i})^{\beta_i}-log\frac{\beta_i}{\alpha_i} + log\Gamma(\frac{1}{\beta_i})
\end{aligned}
\end{equation}

\section{Experimental}
This section describes the experimental environment, setup details, and the datasets used in the experiments. To showcase the superior performance of the proposed UMFFNet for image deraining, we performed comparisons with state-of-the-art methods using both synthetic and real datasets. Additionally, comprehensive ablation experiments were conducted to show the importance of each component in the proposed network.

\begin{figure*}[htbp]
\centering
\includegraphics[scale=1.2]{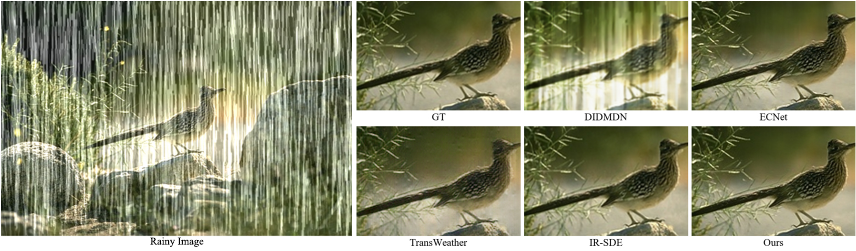}
\caption{Visual comparison of deraining results from Rain100H dataset.}
\label{fig_5}
\end{figure*}

\begin{figure*}[htbp]
\centering
\includegraphics[scale=1.2]{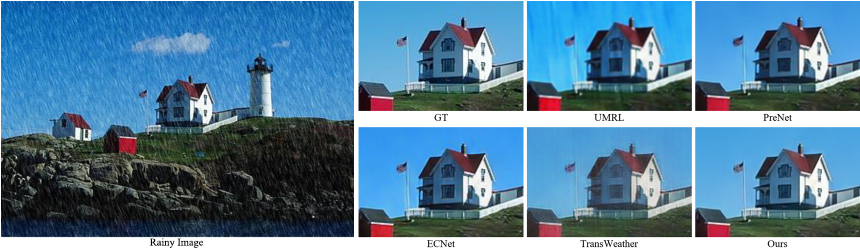}
\caption{Visual comparison of deraining results from Rain800 dataset.}
\label{fig_6}
\end{figure*}

\begin{figure*}[htbp]
\centering
\includegraphics[scale=1.2]{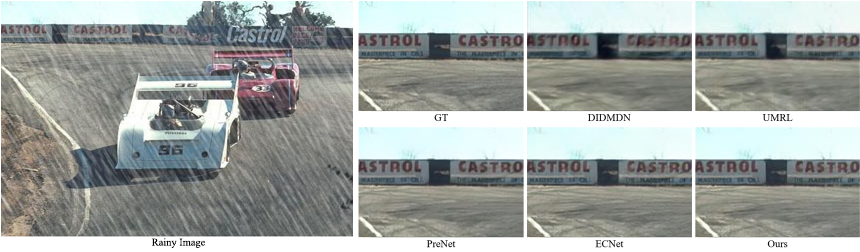}
\caption{Visual comparison of deraining results from Rain1400 dataset.}
\label{fig_7}
\end{figure*}

\subsection{Experimental Settings}
\textbf{Implementation Details.}
The experiments were performed using an NVIDIA RTX 3090 GPU for network training. The network underwent a complete training process for 300 epochs, starting with an initial learning rate of $10^{-3}$. And the learning rate was reduced by a factor of 0.5 every 50 epochs. The batch size was set to 8. Data augmentation involved random cropping of images to $256\times256$, and each patch had a 50\% probability of being horizontally flipped.

For simplicity, we set the number of Residual Attention Blocks at each level as \{N, N, 2N, N, N\}, and the channel numbers as \{C, 2C, 4C, 2C, C\}, where N represents the base block number and C represents the base channel number. To evaluate the scalability of UMFFNet, we introduced three variants of UMFFNet (-T, -B, -L denoting tiny, basic, and large, respectively). All variants were assigned the same channel number, C=32. The three variants varied in their depths, with the base block number N set to \{1, 10, 20\} for each variant, respectively.

\begin{table}[htbp]
\caption{ Descriptions of different synthetic and real-world datasets.}
\centering
\footnotesize
\begin{tabular}{ c| c| c |c |c| c}
   \toprule
       Datasets & Rain100L& Rain100H &  Rain800 & Rain1400 & Fu et al.\\
   \midrule
   Train-Set & 1800 & 1800 & 800& 12600 & - \\
   Test-Set & 100 & 100 & 100 & 1400& 300 \\
   \midrule 
   Type & \multicolumn{4}{c|}{Synthetic} & {Real-world}\\
   \bottomrule
\end{tabular}
	\label{tab_1}
\end{table}

\begin{table*}[htbp]
\caption{Averaged PNSR, SSIM with state-of-the-art deraining algorithms on four synthetic benchmark datasets. The $1^{st}$, $2^{nd}$ and $3^{rd}$ ranks are shown in \textcolor{red}{ red }, \textcolor{blue}{blue} and \textcolor{green}{green}, respectively.}
	\label{tab_2}
\centering
\footnotesize
\begin{tabular}{*{10}{c}}
  \toprule
  \multirow{2}*{Method} & \multirow{2}*{Publication} & \multicolumn{2}{c}{Rain100L} & \multicolumn{2}{c}{Rain100H}  & \multicolumn{2}{c}{Rain800} & \multicolumn{2}{c}{Rain1400} \\
  \cmidrule(lr){3-4}\cmidrule(lr){5-6}\cmidrule(lr){7-8}\cmidrule(lr){9-10}
  & & PSNR↑ & SSIM↑ & PSNR↑ & SSIM↑ & PSNR↑ & SSIM↑ & PSNR↑ & SSIM↑   \\
  \midrule
  RESCAN & CVPR'16 & 29.80 & 0.881 & 26.36 & 0.786 & 25.00 & 0.835 & 31.29 & 0.904 \\
  DIDMDN & CVPR'18 & 25.23  & 0.741 & 17.35 & 0.524& 22.26 & 0.818 & 28.13 & 0.867 \\
  UMRL & CVPR'19 & 29.18 & 0.923 & 26.01& 0.832 & 24.41 & 0.829 & 29.97 & 0.905 \\
  PreNet & CVPR'19 & 32.44 & 0.950 & 26.77 & 0.858 & 24.81 & 0.851 &31.75 & 0.916\\
  SPANet & CVPR'19 & 35.33 & 0.970 & 25.11 & 0.833 & 24.37 & 0.861 &28.57 & 0.891\\
  LPNet & TNNLS'19 & 33.39 & 0.958 & 24.39 & 0.820 & 25.26 & 0.781 & 22.03 & 0.800\\
  HiNet & CVPRW'21 & 37.28 & 0.970 & 30.65 & 0.894 & \textcolor{green}{28.01} & \textcolor{green}{0.870} & \textcolor{red}{33.77} & \textcolor{red}{0.939}\\
  ECNet & WACV'22 & 38.21 & 0.981 & 29.80 & 0.903 & \textcolor{red}{28.80} & \textcolor{red}{0.901} & \textcolor{blue}{32.39} & \textcolor{blue}{0.933}\\
  DSM-Net & CVPR'22 & 38.27 & 0.982 & 28.62 & 0.902 & 27.76 & 0.871 & 30.93 & 0.929 \\
  TransWeather & CVPR'22 & 31.63 & 0.938 & 25.09 & 0.780 & 23.93 & 0.790 &30.19 &0.901  \\
  IR-SDE & ICML'23 & 38.30 & 0.981 & \textcolor{green}{31.65} & \textcolor{green}{0.904} & -& - & - & - \\
  UMFFNet-T & - & \textcolor{green}{38.86} & \textcolor{green}{0.983} &29.65 & 0.889 & 26.36 & 0.833 & 31.04 & 0.915 \\
  UMFFNet-B & - & \textcolor{blue}{41.20} & \textcolor{blue}{0.988} & \textcolor{blue}{32.80} & \textcolor{blue}{0.929} & 27.31 & 0.864 & 31.67 & 0.921\\
  UMFFNet-L & - & \textcolor{red}{41.63} & \textcolor{red}{0.989} & \textcolor{red}{33.46} &  \textcolor{red}{0.935} &  \textcolor{blue}{28.77} &  \textcolor{blue}{0.880} &  \textcolor{green}{31.92} & \textcolor{green}{0.923}\\
  \bottomrule
\end{tabular}
\end{table*}

\textbf{Datasets.}
We utilized four synthetic datasets, including Rain100L\cite{yang2017deep}, Rain100H\cite{yang2017deep}, Rain800\cite{zhang2019image}, and Rain1400\cite{fu2017removing}, as well as a real dataset proposed by Fu et al.\cite{fu2019lightweight}, to evaluate the deraining performance of the proposed method. These datasets encompass rain streaks of varying sizes, densities, and orientations. The detailed composition of the datasets is presented in Tab. \ref{tab_1}.

\subsection{ Comparison with State-of-the-arts}
\textbf{Results on Synthetic Datasets.}
Our proposed UMFFNet was evaluated on four benchmark datasets: Rain100H, Rain100L, Rain800, and Rain1400. We compared it with eleven state-of-the-art deraining methods, including  RESCAN\cite{li2016rain}, DIDMDN\cite{zhang2018density}, UMRL\cite{yasarla2019uncertainty}, PreNet\cite{ren2019progressive}, SPANet\cite{wang2019spatial}, LPNet\cite{fu2019lightweight}, HiNet\cite{chen2021hinet}, ECNet\cite{li2022single}, DSM-Net\cite{li2022deep}, TransWeather\cite{valanarasu2022transweather} and IR-SDE\cite{luo2023image}.

Tab. \ref{tab_2} displays the objective results of all tested algorithms on the synthetic datasets. Our UMFFNet outperformed other methods with significant improvements in terms of PSNR and SSIM on the Rain100H and Rain100L datasets. We also observed that our UMFFNet-T, with minimal computational cost, outperformed other state-of-the-art methods on the Rain100L dataset.

In Fig. \ref{fig_5}-\ref{fig_7}, we present visual results of different deraining methods applied to various synthetic datasets, displaying magnified regions of selected images. Upon observing the details in these magnified images, it becomes apparent that while the compared methods successfully eliminate a significant portion of rain streaks, they also introduce varying degrees of background blurring, loss of original background edge information, and even unnecessary dark noise in the images. In contrast, UMFFNet effectively preserves the texture and edge details of the image while removing rain streaks by adequately leveraging multi-scale feature information and uncertainty information. This visually appealing outcome enhances the overall visual quality of the images.

\begin{figure*}[htbp]
\centering
\includegraphics[scale=1.2]{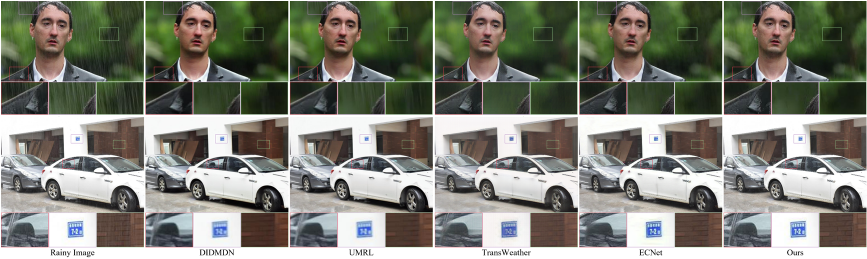}
\caption{Visual comparison of deraining results from real-world rainy images.}
\label{fig_8}
\end{figure*}

\textbf{Results on Real-world Images.}
To validate the effectiveness of the proposed method in practical applications, we compared it with other methods on a real-world dataset. To ensure fairness, all methods utilized pre-trained models trained on the Rain100H dataset for real-world image deraining. As depicted in Fig. \ref{fig_8}, from the magnified region images, both DIDMDN and UMRL not only fail to completely eliminate rain streaks but also blur the texture edges of objects. Although TransWeather and ECNet can effectively remove rain streaks, they encounter difficulties in preserving the texture details of the background objects. Overall, our UMFFNet generates deraining results that are more natural and satisfactory compared to other methods.

\begin{table*}[htbp]
\caption{ Ablation Analysis on UMFFNet.}
\centering
\footnotesize
\begin{tabular}{ c c c c c c c}
   \toprule
       Variants & Base & $V_1$ &  $V_2$ & $V_3$ & $V_4$ & $V_5$\\
   \midrule
   Uncertainty Estimation & w/o & \checkmark & \checkmark & \checkmark& \checkmark & \checkmark \\ 
   UFFB & w/o & w/o & \checkmark & \checkmark& \checkmark & \checkmark \\
   SFFB & w/o & w/o & w/o & \checkmark & \checkmark & \checkmark\\
   MFB & w/o & w/o & w/o & w/o & \checkmark & \checkmark\\
   RAB & w/o & w/o & w/o & w/o & w/o & \checkmark\\
   \midrule
   PSNR & 34.67 & 40.45 & 40.76 & 41.04 & 39.57 & \textbf{41.63}\\ 
   SSIM & 0.926 & 0.980 & 0.981 & 0.981 & 0.986 & \textbf{0.989}\\ 
   \bottomrule
\end{tabular}
	\label{tab_3}
\end{table*}

\subsection{ Ablation Study}
\textbf{The effect of the proposed components.}
Compared to other state-of-the-art methods, our UMFFNet demonstrates superior deraining performance. To further investigate the effectiveness and rationality of different components within UMFFNet, we conducted comprehensive ablation experiments. All the ablation experiments were performed on the Rain100L dataset while ensuring consistent parameter settings.

Initially, we built a base network by utilizing the original multi-input/output UNet architecture. The base network only incorporated the encoder-decoder part of the proposed network, while utilizing ResNet blocks as the base blocks. Subsequently, we progressively added or replaced different components within the network, as outlined below:
\begin{enumerate}
\item base network + Uncertainty Estimation $\rightarrow$ $V_1$,
\item $V_1$ + Uncertainty Feature Fusion Block(UFFB) $\rightarrow$ $V_2$, 
\item $V_2$ + Supervised Feature Fusion Block(SFFB) $\rightarrow$ $V_3$, 
\item $V_3$ + Multi-stage Fusion Block(MFB) $\rightarrow$ $V_4$, 
\item $V_4$ + Residual Attention Block(RAB) $\rightarrow$ $V_5$(full model),
\end{enumerate}
All these variants were retrained using the same approach. The experimental results are presented in Tab. \ref{tab_3}, revealing that solely estimating the output uncertainty does not effectively enhance the network's performance. However, the combination of the Uncertainty Feature Fusion Block with uncertainty estimation significantly improves the deraining performance of the network. Furthermore, the inclusion of the SFFB and MFB aids in more effectively expanding the fusion channels of feature information, preventing feature information loss, and thus contributing significantly to the performance improvement of the network. Lastly, the addition of the simplified attention mechanism enables the network to improve its feature learning capability at a reduced computational cost, resulting in further enhancement of derain performance.

\begin{figure*}[htbp]
\centering
\includegraphics[scale=1.0]{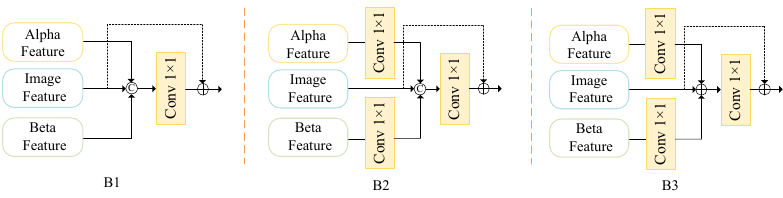}
\caption{The three kinds of blocks with different structures based UFFB.}
\label{fig_9}
\end{figure*}

\begin{table}[htbp]
\caption{ Effects of different structures of the proposed UFFB.}
\centering
\footnotesize
\begin{tabular}{ c c c c}
   \toprule
       Experiments & B1 & B2 &  B3\\
   \midrule
   Conv layer (before fusion) &  & \checkmark & \checkmark \\ 
   Sum. up &  &  & \checkmark \\
   Concat. & \checkmark & \checkmark &  \\
   \midrule
   PSNR & 39.84 & 40.96 & \textbf{41.63} \\ 
   SSIM & 0.986 & 0.988 & \textbf{0.989} \\ 
   \bottomrule
\end{tabular}
	\label{tab_4}
\end{table}

\textbf{The effect of the proposed UFFB.}
Due to the substantial contribution of UFFB to network performance, it is crucial to analyze its various structures. Fig. \ref{fig_9} illustrates other alternative structures. The results, as shown in Tab. \ref{tab_4}, indicate that the block employed in this paper outperforms other blocks in terms of performance.

\begin{table}[htbp]
\caption{ Parameters and Flops of different methods. Our UMFFNet achieves the best output performance with appropriate Params and Flops.}
	\label{tab_5}
\centering
\footnotesize
\begin{tabular}{ c c c c}
   \toprule
       Method   & Params(M) & Flops(G) &  PSNR/SSIM  \\
   \midrule
   UMRL & 0.98 & 16.55 & 29.18/0.923 \\
   HiNet & 88.67 & 170.73 & 37.28/0.970 \\
   MPRNet & 20.1 & 15.84 & 36.40/0.965\\
   TransWeather & 38.05 & 6.12 & 31.63/0.938\\
   ECNet & 2.46 & 15.84 & 38.21/0.981\\
   UMFFNet-T & 1.52 & 17.33 & 38.83/0.983\\
   UMFFNet-B & 8.88 & 82.89 & 41.20/0.988\\
   UMFFNet-L & 17.07 & 155.66 & 41.63/0.989 \\
   \bottomrule
\end{tabular}
\end{table}

\subsection{ Computational Costs and Time}
We compared the computational cost and runtime of the proposed UMFFNet with other state-of-the-art methods. Each test was conducted under the same environment, utilizing an NVIDIA GeForce RTX 3090 GPU as the running device.

To fairly represent the computational cost of different methods, we calculated the parameters and floating-point operations per second (Flops) by providing a $3\times256\times256$ image as input. The results, as shown in Tab. \ref{tab_5}, the PSNR/SSIM values are the deraining results of different methods on the Rain100L dataset. In terms of parameters, our UMFFNet-T is remarkably lightweight. Compared to HiNet, UMFFNet-T achieves superior deraining performance with only approximately 1.75\% of the parameters and 10.2\% of the Flops. Furthermore, UMFFNet-L also achieves the best deraining performance with approximately 19.3\% of the parameters.

Tab. \ref{tab_6} displays the comparison results of different deraining methods in terms of runtime on the Rain100L dataset. Our UMFFNet-T and ECNet exhibit the highest runtime efficiency, with UMFFNet-T achieving better deraining performance compared to other competing methods. On the other hand, UMFFNet-L achieves the best PSNR/SSIM results with slightly higher runtime than UMFFNet-T.

\begin{table}[htbp]
\caption{ Average computational time (in seconds) of different methods evaluated on the Rain100L dataset.}
	\label{tab_6}
\centering
\footnotesize
\begin{tabular}{ c c c}
   \toprule
       Method & Platform & Average time\\
   \midrule
   DIDMDN & PyTorch(GPU) & 0.108\\
   UMRL & PyTorch(GPU) & 5.880 \\
   ECNet & PyTorch(GPU) & \textbf{0.020}\\
   TransWeather & PyTorch(GPU) & 0.152 \\
   IR-SDE & PyTorch(GPU) & 14.010 \\
   UMFFNet-T & PyTorch(GPU) & \textbf{0.020} \\
   UMFFNet-B & PyTorch(GPU) & 0.031\\
   UMFFNet-L & PyTorch(GPU) & 0.043 \\
   \bottomrule
\end{tabular}
\end{table}

\section{Conclusion}
This paper proposes an uncertainty-driven multi-scale feature fusion network, UMFFNet, for obtaining reliable and clean deraining results. Specifically, to fully utilize the feature correlations of multi-scale rain streaks and achieve high-quality deraining, we design several foundational blocks (RAB, SFFB, MFB) to ensure collaborative interactions across multiple levels, expanding the fusion channels of feature information and avoiding information loss. Additionally, we model the pixel residuals with a generalized Gaussian distribution to predict the uncertainty of the outputs. Leveraging the estimated uncertainty maps, we develop an uncertainty feature fusion block (UFFB) to adaptively improve the representation of uncertain pixels in regions with significant reconstruction errors. Experiments demonstrate significant performance improvements of our model over benchmark datasets, surpassing the state-of-the-arts. 

Notably, our model achieves efficient and superior deraining tasks with minimal parameter cost, which is particularly meaningful for resource-constrained devices. Moreover, our model, equipped with well-calibrated uncertainty, gains a significant competitive advantage in measurement systems.


%

\bibliographystyle{unsrt}
\bibliography{reference.bib}

\begin{IEEEbiography}[{\includegraphics[width=1in,height=1.25in,clip,keepaspectratio]{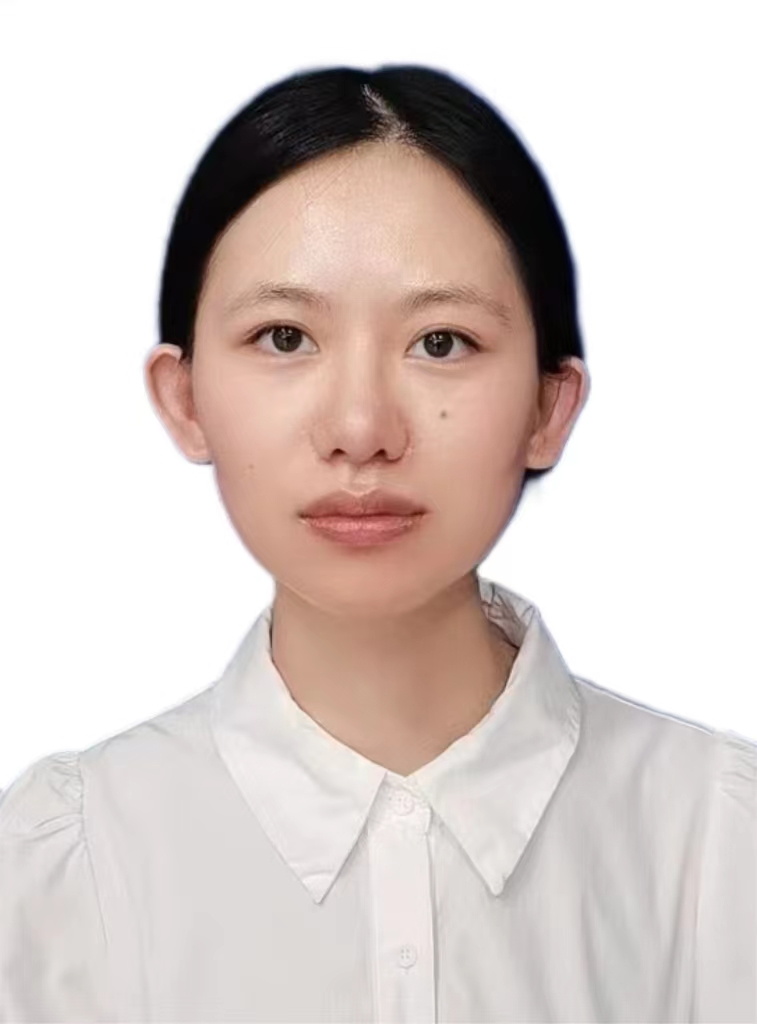}}]{Ming Tong} received the B.E. degree from the Sichuan Agricultural University, Yaan, China, in 2022. She is currently pursuing the M.S. degree with the Nanjing University of Aeronautics and Astronautics (NUAA), Nanjing, China. Her research interests include deep learning, image deraining, object detection, and computer vision.
\end{IEEEbiography}

\begin{IEEEbiography}[{\includegraphics[width=1in,height=1.25in,clip,keepaspectratio]{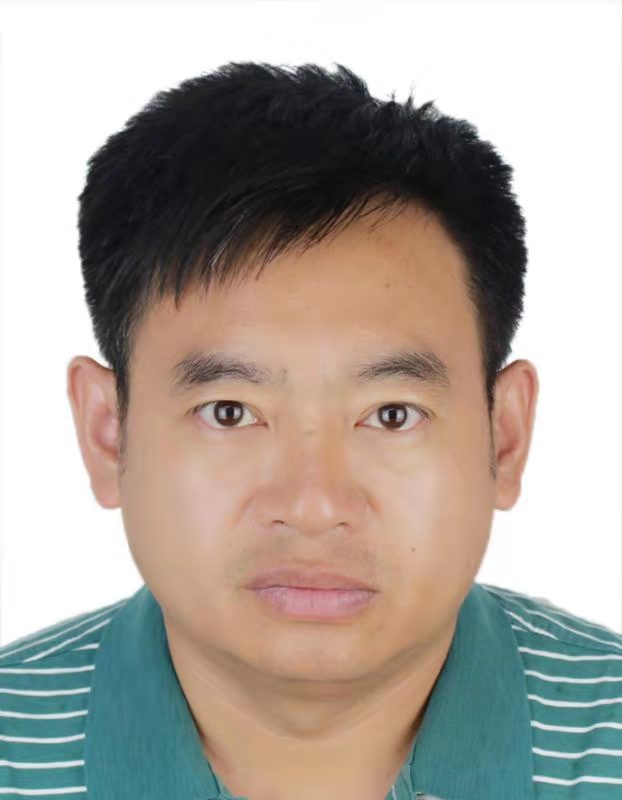}}]{Xuefeng Yan}
	is a professor of the School of Computer Science and Technology, Nanjing University of Aeronautics and Astronautics (NUAA), China. He obtained his PhD degree from Beijing Institute of Technology in 2005. He was the visiting scholar at Georgia State University in 2008 and 2012. His research interests include intelligent computing, MBSE/complex system modeling, simulation and evaluation.
\end{IEEEbiography}

\begin{IEEEbiography}[{\includegraphics[width=1in,height=1.25in,clip,keepaspectratio]{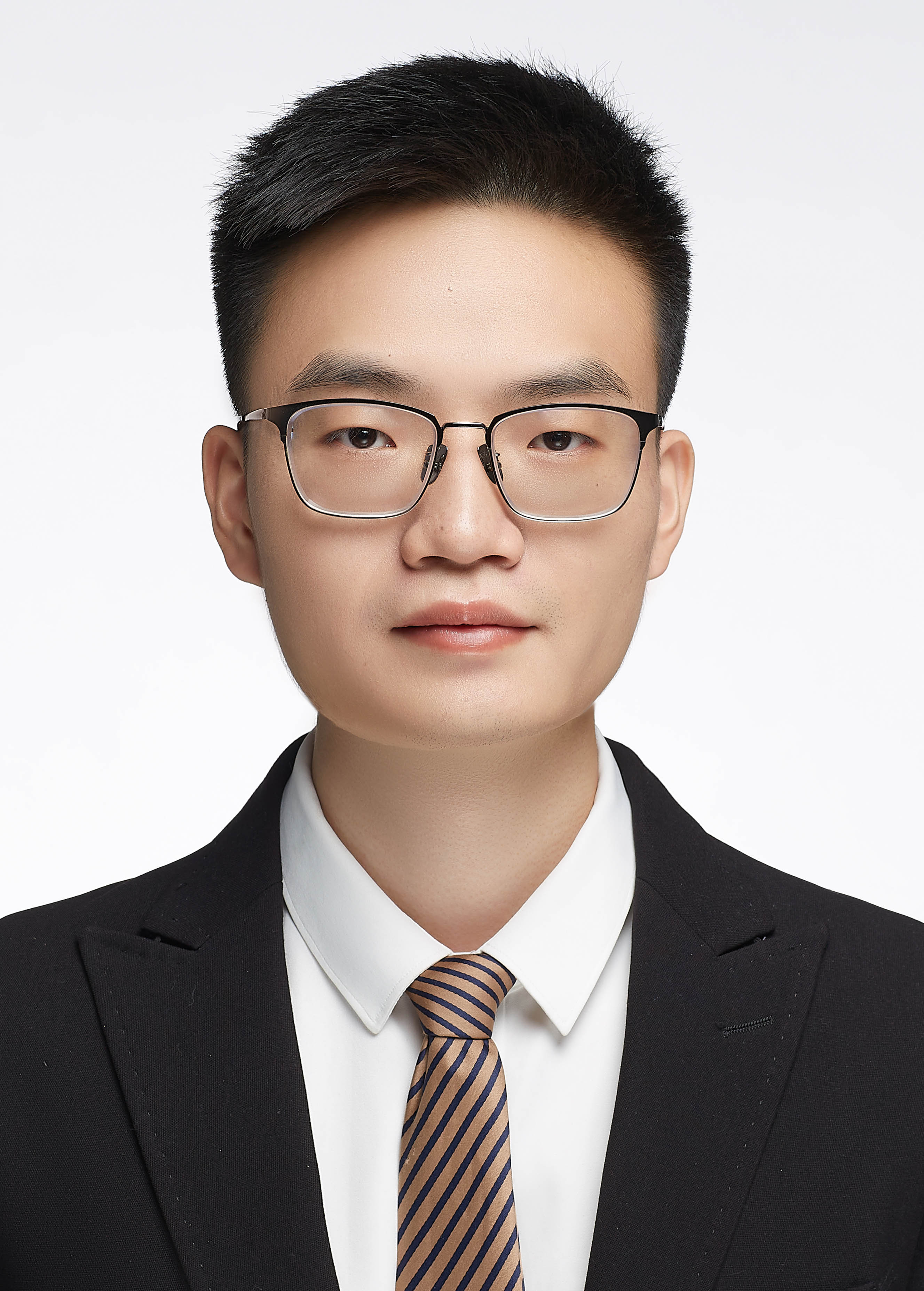}}]{Yongzhen Wang}
received the M.S. degree in 2019. He is currently pursuing the Ph.D. degree with the School of Computer Science and Technology, Nanjing University of Aeronautics and Astronautics (NUAA), Nanjing, China. He has published more than 10 research publications, including IEEE Transactions on Intelligent Transportation Systems, IEEE Transactions on Geoscience and Remote Sensing, Expert Systems with Applications, Computer Graphics Forum, etc. His research interests include deep learning, image processing, and computer vision, particularly in the domains of image dehazing and image deraining. He has served as a PC member for AAAI 2022 and 2023.
\end{IEEEbiography}


 




\vfill

\end{document}